\documentclass{article}
\usepackage{ijcai18}

\usepackage{times}
\usepackage{xcolor}
\usepackage{soul}
\usepackage[utf8]{inputenc}
\usepackage[small]{caption}
\usepackage{helvet}
\usepackage{courier}
\usepackage{amsfonts}
\usepackage{amssymb}
\usepackage{amsmath,bm}
\usepackage{graphicx}
\usepackage{algorithm}
\usepackage{algorithmic}
\usepackage{multirow}
\usepackage{amsmath}
\usepackage{xcolor}
\usepackage{subfigure}
\usepackage{url}
\usepackage{breqn}
\usepackage{epstopdf}
\usepackage{slashbox}
\usepackage{hyphenat}
\title{Building Ethics into Artificial Intelligence}

\author{
Han Yu$^{1,2,3}$,
Zhiqi Shen$^{1,2,3}$,
Chunyan Miao$^{1,2,3}$,
Cyril Leung$^{2,3,4}$,
Victor R. Lesser$^5$,
Qiang Yang$^6$
\\
$^1$ School of Computer Science and Engineering, Nanyang Technological University\\
$^2$ LILY Research Centre, Nanyang Technological University\\
$^3$ Alibaba-NTU Singapore Joint Research Institute\\
$^4$ Department of Electrical and Computer Engineering, The University of British Columbia\\
$^5$ School of Computer Science, University of Massachusetts Amherst\\
$^6$ Department of Computer Science and Engineering, Hong Kong University of Science and Technology\\
\{han.yu, zqshen, ascymiao\}@ntu.edu.sg, cleung@ece.ubc.ca, lesser@cs.umass.edu, qyang@cse.ust.hk
}

\begin{document}

\maketitle

\begin{abstract}
As artificial intelligence (AI) systems become increasingly ubiquitous, the topic of AI governance for ethical decision-making by AI has captured public imagination. Within the AI research community, this topic remains less familiar to many researchers. In this paper, we complement existing surveys, which largely focused on the psychological, social and legal discussions of the topic, with an analysis of recent advances in technical solutions for AI governance. By reviewing publications in leading AI conferences including AAAI, AAMAS, ECAI and IJCAI, we propose a taxonomy which divides the field into four areas: 1) exploring ethical dilemmas; 2) individual ethical decision frameworks; 3) collective ethical decision frameworks; and 4) ethics in human-AI interactions. We highlight the intuitions and key techniques used in each approach, and discuss promising future research directions towards successful integration of ethical AI systems into human societies.
\end{abstract}

\section{Introduction}
As artificial intelligence (AI) technologies enter many areas of our daily life \cite{Cai-et-al:2014,Shi-et-al:2016,Pan-et-al:2017,Zheng-et-al:2018}, the problem of ethical decision-making, which has long been a grand challenge for AI \cite{Wallach-Allen:2008}, has caught public attention. A major source of public anxiety about AI, which tends to be overreactions \cite{Bryson-Kime:2011}, is related to artificial general intelligence (AGI) \cite{Goertzel-Pennachin:2007} research aiming to develop AI with capabilities matching and eventually exceeding those of humans. A self-aware AGI \cite{Dehaene-et-al:2017} with superhuman capabilities is perceived by many as a source of existential risk to humans. Although we are still decades away from AGI, existing autonomous systems (such as autonomous vehicles) already warrant the AI research community to take a serious look into incorporating ethical considerations into such systems.

According to \cite{Cointe-et-al:2016}, \textit{ethics} is a normative practical philosophical discipline of how one should act towards others. It encompasses three dimensions:
\begin{enumerate}
    \item \textit{Consequentialist ethics}: an agent is ethical if and only if it weighs the consequences of each choice and chooses the option which has the most moral outcomes. It is also known as utilitarian ethics as the resulting decisions often aim to produce the best aggregate consequences.
    \item \textit{Deontological ethics}: an agent is ethical if and only if it respects obligations, duties and rights related to given situations. Agents with deontological ethics (also known as duty ethics or obligation ethics) act in accordance to established social norms.
    \item \textit{Virtue ethics}: an agent is ethical if and only if it acts and thinks according to some moral values (e.g. bravery, justice, etc.). Agents with virtue ethics should exhibit an inner drive to be perceived favourably by others.
\end{enumerate}
\textit{Ethical dilemmas} refer to situations in which any available choice leads to infringing some accepted ethical principle and yet a decision has to be made \cite{Kirkpatrick:2015}.

The AI research community realizes that machine ethics is a determining factor to the extent autonomous systems are permitted to interact with humans. Therefore, research works focusing on technical approaches for enabling these systems to respect the rights of humans and only perform actions that follow acceptable ethical principles have emerged. Nevertheless, this topic remains unfamiliar to many AI practitioners and is in need of an in-depth review. However, existing survey papers on the topic of AI governance mostly focused on the psychological, social and legal aspects of the challenges \cite{Arkin:2016,Etzioni-Etzioni:2017,Pavaloiu-Kose:2017}. They do not shed light on technical solutions to implement ethics in AI systems. The most recent survey on technical approaches for ethical AI decision-making was conducted in 2006 \cite{Mclaren:2006} and only covered single agent decision-making approaches.

\begin{table*}[ht]
\caption{A taxonomy of AI governance techniques.} \label{tb:1}
\centering
\resizebox*{1\textwidth}{!}{
\begin{tabular}{|c|c|c|c|}\hline
Exploring Ethical   & Individual Ethical     & Collective Ethical    & Ethics in Human-AI \\
Dilemmas            & Decision Frameworks    & Decision Frameworks   & Interactions \\\hline
\cite{Anderson-Anderson:2014}       & \cite{Dehghani-et-al:2008}     & \cite{Singh:2014,Singh:2015}  & \cite{Battaglino-Damiano:2015}\\
\cite{Bonnefon-et-al:2016}          & \cite{Blass-Forbus:2015}       & \cite{Pagallo:2016}           & \cite{Stock-et-al:2016}\\
\cite{Sharif-et-al:2017}            & \cite{vanRiemsdijk:2015}       & \cite{Greene-et-al:2016}      & \cite{Luckin:2017}\\
                                    & \cite{Cointe-et-al:2016}       & \cite{Noothigattu-et-al:2018} & \cite{Yu-et-al:2017npj}\\
                                    & \cite{Conitzer-et-al:2017}     &                               & \\
                                    & \cite{Berreby-et-al:2017}      &                               &\\
                                    & \cite{Loreggia-et-al:2018}     &                               &\\
                                    & \cite{Wu-Lin:2018}             &                               &\\\hline
\end{tabular}
}
\end{table*}

In this paper, we survey recent advances in techniques for incorporating ethics into AI to bridge this gap. We focus on recent advances published in leading AI research conferences including AAAI, AAMAS, ECAI and IJCAI, as well as articles from well-known journals. We propose a taxonomy which divides the field into four areas (Table \ref{tb:1}):
\begin{enumerate}
    \item \textit{Exploring Ethical Dilemmas}: technical systems enabling the AI research community to understand human preferences on various ethical dilemmas;
    \item \textit{Individual Ethical Decision Frameworks}: generalizable decision-making mechanisms enabling individual agents to judge the ethics of its own actions and the actions of other agents under given contexts;
    \item \textit{Collective Ethical Decision Frameworks}: generalizable decision-making mechanisms enabling multiple agents to reach a collective decision on the course of action that is ethical; and
    \item \textit{Ethics in Human-AI Interactions}: frameworks that incorporate ethical considerations into agents which are designed to influence human behaviours.
\end{enumerate}
Promising future research directions which may enable ethical AI systems to be successfully integrated into human societies are discussed at the end.

\section{Exploring Ethical Dilemmas}
In order to build AI systems that behave ethically, the first step is to explore the ethical dilemmas in the target application scenarios. Recently, software tools based on expert review and crowdsourcing have emerged to serve this purpose.

In \cite{Anderson-Anderson:2014}, the authors proposed the \textit{GenEth} ethical dilemma analyzer. They realized that ethical issues related to intelligent systems are likely to exceed the grasp of the original system designers, and designed GenEth to include ethicists into the discussion process in order to codify ethical principles in given application domains. The authors proposed a set of representation schemas for framing the discussions on AI ethics. It includes:
\begin{enumerate}
    \item \textit{Features}: denoting the presence or absence of factors (e.g., harm, benefit) with integer values;
    \item \textit{Duties}: denoting the responsibility of an agent to minimize/maximize a given feature;
    \item \textit{Actions}: denoting whether an action satisfies or violates certain duties as an integer tuple;
    \item \textit{Cases}: used to compare pairs of actions on their collective ethical impact; and
    \item \textit{Principles}: denoting the ethical preference among different actions as a tuple of integer tuples.
\end{enumerate}
GenEth provides a graphical user interface for discussing ethical dilemmas in a given scenario, and applies inductive logic programming to infer principles of ethical actions.

Whereas GenEth can be regarded as an ethical dilemma exploration tool based on expert review, the Moral Machine project\footnote{\url{http://moralmachine.mit.edu/}} from Massachusetts Institute of Technology (MIT) leverages the wisdom of the crowd to find resolutions for ethical dilemmas. The Moral Machine project focuses on studying the perception of autonomous vehicles (AVs) which are controlled by AI and has the potential to harm pedestrians and/or passengers if they malfunction. When a human driver who has used due caution encounters an accident, the instinct for self-preservation coupled with limited time for decision-making makes it hard to blame him/her for hurting others on ethical grounds. However, when the role of driving is delegated to an AI system, ethics becomes an unavoidable focal point of AV research since designers have the time to program logics for making decisions under various accident scenarios.

The Moral Machine project allows participants to judge various ethical dilemmas facing AVs which have malfunctioned, and select which outcomes they prefer. Then, the decisions are analyzed according to different considerations including: 1) saving more lives, 2) protecting passengers, 3) upholding the law, 4) avoiding intervention, 5) gender preference, 6) species preference, 7) age preference, and 8) social value preference. The project also provides a user interface for participants to design their own ethical dilemmas to elicit opinions from others.

Based on feedbacks from 3 million participants, the Moral Machine project found that people generally prefer the AV to make sacrifices if more lives can be saved. If an AV can save more pedestrian lives by killing its passenger, more people prefer others' AVs to have this feature rather than their own AVs  \cite{Bonnefon-et-al:2016,Sharif-et-al:2017}. Nevetheless,  self-reported preferences often do not align well with actual behaviours \cite{Zell-Krizan:2014}. Thus, how much the findings reflect actual choices is still an open question. There are also suggestions from others that under such ethical dilemmas, decisions should be made in a random fashion (i.e. let fate decide) possibly based on considerations in \cite{Broome:1984}, while there are also calls for AVs to be segregated from human traffic \cite{Bonnefon-et-al:2016}. Such diverse opinions underscore the challenge of automated decision-making under ethical dilemmas.

\section{Individual Ethical Decision Frameworks}
When it comes to ethical decision-making in AI systems, the AI research community largely agrees that generalized frameworks are preferred over ad-hoc rules. Flexible incorporation of norms into AI to enable ethical user and prevent unethical use is useful since ethical bounds can be contextual and difficult to define as design time. Nevertheless, if updates are provided by people, some review mechanisms should be put in place to prevent abuse \cite{vanRiemsdijk:2015}.

In \cite{Dehghani-et-al:2008}, the authors observed that moral decision-making by humans not only involves utilitarian considerations, but also moral rules. These rules are acquired from past example cases and are often culturally sensitive. Such rules often involve \textit{protected values} (a.k.a. \textit{sacred values}), which morally forbids the commitment of certain actions regardless of consequences (e.g., the act of attempting to murder is morally unacceptable regardless the outcome). The authors proposed \textit{MoralDM} which enables an agent to resolve ethical dilemmas by leveraging on two mechanisms: 1) first-principles reasoning, which makes decisions based on well-established ethical rules (e.g., protected values); and 2) analogical reasoning, which compares a given scenario to past resolved similar cases to aid decision-making. As the number of resolved cases increases, the exhaustive comparison approach by MoralDM is expected to become computationally intractable. Thus, in \cite{Blass-Forbus:2015}, MoralDM is extended with structure mapping which trims the search space by computing the correspondences, candidate inferences and similarity scores between cases to improve the efficiency of analogical generalization.

A framework that enables agents to make judgements on the ethics of its own and other agents' actions was proposed in \cite{Cointe-et-al:2016}. It contains representations of ethics based on theories of good and theories of right, and ethical judgement processes based on awareness and evaluation. The proposed agent ethical judgement process is based on the Belief-Desire-Intention (BDI) agent mental model \cite{Rao-Georgeff:1995}. To judge the ethics of an agent's own actions, the awareness process generates the beliefs that describe the current situation facing the agent and the goals of the agent. Based on the beliefs and goals, the evaluation process generates the set of possible actions and desirable actions. The goodness process then computes the set of ethical actions based on the agent's beliefs, desires, actions, and moral value rules. Finally, the rightness process evaluates whether or not executing a possible action is right under the current situation and selects an action which satisfies the rightfulness requirement. When making ethical judgements on other agents, this process is further adapted to the conditions of: 1) blind ethical judgement (the given agent's state and knowledge are unknown); 2) partially informed ethical judgement (with some information about a given agent's state and knowledge); and 3) fully informed ethical judgement (with complete information about a given agent's state and knowledge). Nevertheless, the current framework has no quantitative measure of how far a behaviour is from rightfulness or goodness.

In \cite{Conitzer-et-al:2017}, the authors proposed two possible ways towards developing a general ethical decision-making framework for AI based on game theory and machine learning, respectively. For the game theory based framework, the authors suggest the extensive form (a generalization of game trees) as a foundation scheme to represent dilemmas. As the current extensive form does not account for protected values in which an action can be treated as unethical regardless of its consequence, the authors proposed to extend the extensive form representation with passive actions for agents to select in order to be ethical. For machine learning based ethical decision-making, the key approach is to classify whether a given action under a given scenario is morally right or wrong. In order to achieve this goal, well-labeled training data, possibly from human judgements, should be acquired. The Moral Machine project mentioned in the previous section could be a possible source of such data, although we may have to take into account potential inconsistencies as a result of cultural backgrounds and other factors before using such data for training. The main challenge in machine learning based moral decision-making is to design a generalizable representation of ethical dilemmas. Existing approaches which identify nuanced features based on insights into particular application scenarios may not be enough for this purpose. The authors suggest leveraging psychological frameworks of moral foundation (e.g., harm/care, fairness/reciprocity, loyalty, authority and purity) \cite{Clifford-et-al:2015} as bases for developing a generalizable representation of ethical dilemmas for machine learning-based approaches. Game theory and machine learning can be combined into one framework in which game theoretic analysis of ethics is used as a feature to train machine learning approaches, while machine learning helps game theory identify ethical aspects which are overlooked.

Ethics requirements are often exogenous to AI agents. Thus, there needs to be some ways to reconcile ethics requirements with the agents' endogenous subjective preferences in order to make ethically aligned decisions. In \cite{Loreggia-et-al:2018}, the authors proposed an approach to leverage the CP-net formalism to represent the exogenous ethics priorities and endogenous subjective preferences. The authors further established a notion of distance between CP-nets so as to enable AI agents to make decisions using their subjective preferences if they are close enough to the ethical principles. This approach helps AI agents balance between fulfilling their preferences and following ethical requirements.

So far, the decision-making frameworks with ethical and moral considerations reviewed put the burden of codifying ethics on AI system developers. The information on what is morally right or wrong has to be incorporated into the AI engine during the development phase. In \cite{Berreby-et-al:2017}, the authors proposed a high level action language for designing ethical agents in an attempt to shift the burden of moral reasoning to the autonomous agents. The framework collects action, event and situation information to enable an agent to simulate the outcome of various courses of actions. The event traces are then passed to the causal engine to produce causal traces. The ethical specifications and priority of ethical considerations under a given situation are used to compute the goodness assessment on the consequences. These outputs are then combined with deontological specifications (duties, obligations, rights) to produce a final rightfulness assessment. The framework is implemented with answer set programming \cite{Lifschitz:2008}. It has been shown to be able to generate rules to enable agents to decide and explain their actions, and reason about other agents' actions on ethical grounds.

Reinforcement learning (RL) \cite{Sutton-Barto:1998} is one of the commonly used decision-making mechanisms in AI. In \cite{Wu-Lin:2018}, the authors investigated how to enable RL to take ethics into account. Leveraging on the well-established technique of reward shaping in RL which incorporates prior knowledge into the reward function to speed up the learning process, the authors proposed the \textit{ethics shaping} approach to incorporate ethical values into RL. By assuming that the majority of observed human behaviours are ethical, the proposed approach learns ethical shaping policies from available human behaviour data in given application domains. The ethics shaping function rewards positive ethical decisions, punishes negative ethical decisions, and remains neutral when ethical considerations are not involved. Similar in spirit to \cite{Berreby-et-al:2017}, by separating ethics shaping from the RL reward function design, the proposed approach aims to shift the burden of codifying ethics away from RL designers so that they do not need to be well-versed in ethical decision-making in order to develop ethical RL systems.

\section{Collective Ethical Decision Frameworks}
By enabling individual agents to behave ethically and judge the ethics of other agents' actions, is it enough to create a society of well coordinated and collaborative agents acting with human wellbeing as their primary concern? In \cite{Pagallo:2016}, the author believes that this is not enough. The author advocates the need of primary rules governing social norms and allowing the creation, modification and suppression of the primary rules with secondary rules as situations evolve. In this section, we focus on decision-making frameworks which help a collective of autonomous entities (including agents and humans) to select ethical actions together.

In \cite{Singh:2014,Singh:2015}, the author proposed a framework that uses social norms to govern autonomous entities' (e.g., AI agents' or human beings') behaviours. Such an approach is inherently distributed rather than relying on a central authority. Individuals maintain their autonomy through executing their own decision policies, but are subjected to social norms defined by the collective through roles (which require qualifications from individuals, grant them privileges, and impose penalties if they misbehave). Social norms are defined through a template containing codified commitment, authorization, prohibition, sanction and power. The individuals then form a network of trust based on techniques from the reputation modelling literature \cite{Yu-et-al:2010PIEEE,Yu-et-al:2013Access} to achieve collective self-governance through dynamic interactions.

In \cite{Greene-et-al:2016}, the authors envisioned a possible way forward to enable human-agent collectives \cite{Jennings-et-al:2014} to make ethical collective decisions. By imbuing individual agents with ethical decision-making mechanisms (such as those mentioned in the previous section), a population of agents can take on different roles when evaluating choices of action with moral considerations in a given scenario. For instance, some agents may evaluate deontological ethics. Others may evaluate consequentialist ethics and virtue ethics. Based on a set of initial ethics rules, more complex rules can be acquired gradually through learning. Their evaluations, manifested in the form of preferences and limited by feasibility constraints, can be aggregated to reach a collective decision on the choices of actions by leveraging advances in the preference aggregation and multi-agent voting literature.

Nevertheless, the authors of \cite{Greene-et-al:2016} also highlighted the need for new forms of preference representation in collective ethical decision-making. When dealing with ethical decision-making, the potential candidate actions to choose from can vastly outnumber the number of agents involved which is very different from multi-agent voting scenarios. Moreover, the candidate actions may not be independent from each other, some of them may share certain features which describe their ethical dilemma situations. Preference information by agents on actions may be missing or imprecise which introduces uncertainty into the decision-making process. These challenges need to be resolved towards collective ethical decision-making with AI.

Following up on this vision, \cite{Noothigattu-et-al:2018} proposed a voting-based system for autonomous entities to make collective ethical decisions. The proposed approach leverages data collected from the Moral Machine project. Self-reported preference over different outcomes under diverse ethical dilemmas are used to learn models of preference for the human voters over different alternative outcomes. These individual models are then summarized to form a model that approximates the collective preference of all voters. The authors introduced the concept of \textit{swap-dominance}\footnote{Assuming everything else is fixed, an outcome $a$ swap-dominates another outcome $b$ if every ranking which ranks $a$ higher than $b$ has a weight which is equal to or larger than rankings that rank $b$ higher than $a$.} when ranking alternatives to form a model of ethical preferences. When new decisions need to be made, the summarized model is used to compute a collective decision that results in the best possible outcome (i.e. satisfying consequentialist ethics). This is made computationally efficient with the swap-dominance property.

\section{Ethics in Human-AI Interactions}
In AI applications which attempt to influence people's behaviours, the principles established by the Belmont Report \cite{Belmont:1978} for behavioural sciences have been suggested to be a starting point for ensuring ethics \cite{Luckin:2017,Yu-et-al:2017npj}. The principles include three key requirements: 1) people's personal autonomy should not be violated (they should be able to maintain their free will when interacting with the technology); 2) benefits brought about by the technology should outweigh risks; and 3) the benefits and risks should be distributed fairly among the users (people should not be discriminated based on their personal backgrounds such as race, gender and religion). The challenge of measuring benefits and risks remains open for application designers albeit the Ethically Aligned Design guidelines from the IEEE can be a useful starting point \cite{IEEE:2018}. Computational formulations of human centric values (e.g., collective wellbeing and work-life balance) have been proposed and incorporated into the objective functions of recent AI-powered algorithmic management approaches in crowdsourcing \cite{Yu-et-al:2016SciRep,Yu-et-al:2017SciRep,Yu-et-al:2017ICA}.

One of the application areas in which AI attempts to influence people's behaviours is persuasion agents \cite{Kang-et-al:2015,Rosenfeld-Kraus:2016}. In \cite{Stock-et-al:2016}, the authors conducted a large-scale study to investigate human perceptions on the ethics of persuasion by an AI agent. The ethical dilemma used is the trolley scenario which involves making a participant actively cause harm to an innocent bystander by pushing him on to the train track in order to save the lives of five people. It is a consequentialist ethical outcome which requires the decision-maker to violate a sacred value (i.e. one shall not kill). The authors tested three persuasive strategies: 1) appealing to the participants emotionally; 2) presenting the participants with utilitarian arguments; and 3) lying. The three strategies are delivered to some participants by a person playing the role of an authority (the station master of the train station) and by a persuasion agent. The results suggested that participants hold a strong preconceived negative attitude towards the persuasion agent, and argumentation-based and lying-based persuasion strategies work better than emotional persuasion strategies. The findings did not show significant variation across genders or cultures. The study suggests that the adoption of persuasion strategies should take into account differences in individual personality, ethical attitude and expertise in the given domain.

Although emotional appeals may not be an effective persuasive strategy under ethical dilemmas, ethically appropriate emotional responses from agents can enhance human-AI interaction. In \cite{Battaglino-Damiano:2015}, an approach based on the \textit{Coping Theory} \cite{Marsella-Gratch:2003} to allow agents to deal with strong negative emotions by changing the appraisal of the given situation was proposed. The agent assesses the ethical effects of its own actions and other agents' actions. If its own action violates a given moral value, the \textit{shame} emotion is triggered which serves to lower the priority of continuing with the given action. If another agent's action violates a given moral value, the \textit{reproach} emotion is triggered in the observing agent which serves to increase social distance with the given agent (e.g., by reducing trust). The ethical decision-making process is similar to existing individual ethical decision frameworks. The triggering of emotional responses serves as an implicit reward for the agent and facilitates communications with humans in the loop.

\section{Discussions}
Based on recent advances in AI governance techniques, it appears that most work focused on developing generalizable individual ethical decision frameworks combining rule-based and example-based approaches to resolving ethical dilemmas. In order to learn appropriate rules from examples of ethical decision-making by humans, more work on collecting data about various ethical dilemmas from people with different cultural backgrounds is required. Works on collective ethical decision-making based on multi-agent voting have also appeared, but much work is still needed to design mechanisms to represent ethical preferences by agents. How AI can act ethically when making recommendations to humans and express their ethical judgements affectively are the current foci of ethical human-AI interaction research. In addition, AI engineers need to engage more with the ethics and decision making communities. These people want to help and the AI research community be reaching out to them to leverage their expertise in the pursuit of ethical AI technologies. Since such AI technologies as autonomous vehicles, autonomous weapons, and cryptocurrencies are becoming a reality and affecting societies, a global and unified AI regulatory framework needs to be established as soon as possible to address the ethical issues by drawing on interdisciplinary expertise \cite{Erdelyi-Goldsmith:2018}.

In order for ethics to be built into AI, \cite{Burton-et-al:2017,Goldsmith-Burton:2017} advocate that ethics should be part of the AI curricula. This is based on the observation that consequentialist ethics (or ethics based on the utilitarian analysis of possible outcomes) is most closely related to the decision-theoretic frame of mind familiar to today's AI researchers. Deontological ethics (or rule-based ethics) and virtue ethics are less familiar among AI researchers. Understanding deontological ethics can help AI researchers determine which rules are more fundamental and, therefore, should take priority in an ethical decision framework. Understanding virtue ethics, which concerns questions on whom one wishes to become, can help AI researchers frame ethical discussions in the context of changing social conditions (possibly brought on by AI technologies) and guide the incorporation of ethics into AI which shape the paths of learning. Learning materials on these different dimensions of ethics could help AI researchers understand more clearly the topic of ethical decision-making and steer the field of AI towards more emphasis on ethical interactions with humans.

\section{Future Research Directions}
From this survey, we envision several possible future research directions which can impact this field going forward. Firstly, the current mechanism of crowdsourcing self-reported preference on ethical dilemmas as represented by the Moral Machine project has its limitations. Self-reported preferences have been shown to deviate from actual choice behaviours. Researchers from multiple disciplines need to conduct social-systems analysis \cite{Crawford-Calo:2016} of AI in order to understand the impact of AI under different social, cultural and political settings. There may be opportunities for transfer learning \cite{Pan-Yang:2010} to be applied in this case to model different ethics due to diversities in culture and other aspects. The insights from such studies can complement crowdsourced human preference data when building computational models of human ethics. In addition, they may also help AI researchers establish coherent utility functions from apparently inconsistent human ethical judgements.

Secondly, with AI becoming increasingly ubiquitous in our daily life, we may need to consider revising our current social contracts. Research in this area will help us establish regulations about who is responsible when things go wrong with regard to AI, and how to monitor and enforce these regulations. This research direction is inherently dynamic and interdisciplinary in nature as it must be updated with changing cultural, social, legal, philosophical and technological realities.

Thirdly, another important research area for ethical decision-making by AI is to enable AI to explain its decisions under the framework of human ethics. The challenge here is that as deployed AI programs learn to update the decision-making logic, the AI designers may not be able to anticipate all outcomes at design time and may not understand the decisions made by the AI entities later \cite{Venema:2018}. Argumentation-based explainable AI \cite{Fan-Toni:2015,Langley-et-al:2017} can be a good starting point for this purpose as it is well suited to the consequentialist ethics which is a commonly adopted approach for implementing AI ethics. Nevertheless, depending on how the explanations are used, researchers need to strike a balance on the level of details to be included. Full transparency may be too overwhelming if the objective is to persuade a user to follow a time-critical recommendation, but can be useful as a mechanism to trace the AI decision process afterwards. On the other hand, not enough transparency may hamper users' trust in the AI. AI researchers can borrow ideas from the field of mass communication to design proper trade-offs.

Last but not least, the incorporation of ethical considerations into AI systems will influence human-AI interaction dynamics. By knowing that AI decisions follow ethical principles, some people may adapt their behaviours in order to take advantage of this and render the AI systems unable to achieve their design objectives. For example, an ethical autonomous gun system (if there can be such a thing) could be disabled by a child (who is generally regarded as a non-combatant and in need of protection) with a spray paint (which is generally not considered a lethal weapon) painting over the sensor system of the gun. In this case, Adversarial Game Theory \cite{Vorobeychik-et-al:2012} may need to be incorporated into future AI ethical decision frameworks in order to enable AI to preserve the original design objectives in the presence of strategic human behaviours.

\small
\section*{Acknowledgements}
This research is supported by the National Research Foundation, Prime Minister's Office, Singapore under its IDM Futures Funding Initiative; Nanyang Technological University, Nanyang Assistant Professorship (NAP); the Singapore Ministry of Health under its National Innovation Challenge on Active and Confident Ageing (NIC Project No. MOH/NIC/COG04/2017); and the NTU-PKU Joint Research Institute, a collaboration between Nanyang Technological University and Peking University that is sponsored by a donation from the Ng Teng Fong Charitable Foundation.


\bibliographystyle{named}

\end{document}